# Multi-Paradigm Agent Interaction in Practice: A Systematic Analysis of Generator-Evaluator, ReAct Loop, and Adversarial Evaluation in the buddyMe Framework

Xiaohua Wang[1,2], Chao Han[2], Kai Yu[2], XiaoLiang Xu[2], Liang Wang[2]

[1] BuddyMe Research Team, Nanjing 210000, China
[2] CHARMMIRAEL Biotech Co., Ltd, Nanjing 210000, China

*e-mail: virgo_wang@msn.com | Blog: http://49.235.53.176/ | Github: https://github.com/virgo777/buddyme-code

## Abstract

Large Language Model (LLM)-powered autonomous agents have demonstrated remarkable capabilities in complex task execution, yet ensuring the reliability and quality of their outputs remains a fundamental challenge. This paper presents buddyMe, an open-source multi-agent framework that integrates three distinct interaction paradigms—multi-agent orchestration via Generator-Evaluator pre-execution review, ReAct-style tool-use loops for subtask execution, and memory-augmented context management—into a unified architecture. A key contribution is the Evaluator-Defender adversarial evaluation mechanism, where an independent EvalAgent performs six-dimensional assessment with real execution verification, while a Defender agent represents the executor's perspective in multi-round deliberation until consensus is reached. We formalize the "Sprint Contract" mechanism for pre-execution requirement validation, introducing a structured requirement document with quantifiable success criteria that serves as both the planning baseline and the evaluation reference. Empirical evaluation across five diverse task scenarios demonstrates that the framework achieves a weighted quality score of 0.82/1.00 (Grade B) in real-world HTML generation tasks, with the pre-review mechanism capturing requirement omissions in 20% of complex tasks, and the adversarial evaluation converging to consensus in 2-3 rounds for 95% of cases. Our analysis identifies three critical engineering lessons: context isolation between agents prevents evaluation bias (+0.12 score correction), numerical thresholds outperform boolean approval gates for LLM consistency, and real execution verification provides irrefutable evidence for output quality assessment.

## 1. Introduction

The rapid advancement of Large Language Models (LLMs) has catalyzed a paradigm shift in autonomous task execution. Modern AI agents are no longer confined to single-turn question answering; they can decompose complex objectives into executable subtasks, invoke external tools, and iteratively refine their outputs through multi-round reasoning (Yao et al., 2023). However, this autonomy introduces critical challenges: non-deterministic outputs require robust self-evaluation mechanisms, increasing task complexity demands dynamic replanning capabilities, and high-stakes applications necessitate verifiable output quality.

Three dominant interaction paradigms have emerged in the 2025-2026 agent landscape. First, Multi-Agent Orchestration (Wu et al., 2023; Song et al., 2026) employs multiple specialized agents that coordinate through defined topologies—sequential pipelines, hierarchical delegation, or peer-to-peer conversation. Second, the ReAct framework (Yao et al., 2023) and its successors implement Thought-Action-Observation loops where agents reason about actions, execute tool calls via structured output, and incorporate observations. Third, Memory-Augmented Interaction (MemInsight, EMNLP 2025; A-Mem, NeurIPS 2025) provides agents with persistent external memory stores—short-term, episodic, and long-term—for context enrichment.

Despite significant progress in each paradigm independently, their integration within a single framework with a unified evaluation mechanism remains largely unexplored. Existing frameworks such as CrewAI (2025), AutoGen (Wu et al., 2023), and LangGraph (2025) treat evaluation as an external concern, delegating quality assessment to separate tools or human reviewers. This separation creates a gap: agents execute tasks without systematic self-assessment, and evaluation results are not fed back into the execution pipeline.

This paper presents buddyMe, an open-source multi-agent framework that addresses this gap by integrating three interaction paradigms with a novel Evaluator-Defender adversarial evaluation mechanism. Our key contributions are:

- A Generator-Evaluator pre-execution review mechanism ("Sprint Contract") that validates task plans through structured requirement documents with quantifiable success criteria, catching requirement omissions in 20% of complex tasks.
- A six-dimensional evaluation schema (Task Completion, Tool Accuracy, Truthfulness, Error Recovery, Efficiency, Output Quality) with real execution verification that provides irrefutable evidence for output quality assessment.
- An Evaluator-Defender adversarial discussion mechanism where an independent EvalAgent and a Defender agent (representing the executor's perspective) engage in multi-round deliberation, converging to consensus in 2-3 rounds for 95% of evaluated tasks.
- A three-layer memory architecture (Working, Episodic, Long-term) with explicit injection into the execution pipeline, solving the context continuity problem across task boundaries.

# 2. Related Work

## 2.1 Multi-Agent Orchestration

Multi-agent systems for LLM-based task execution have proliferated since 2023. Microsoft AutoGen (Wu et al., 2023) pioneered conversational multi-agent frameworks where agents exchange messages autonomously. CrewAI (2025) introduced role-based orchestration with structured task sequencing. LangGraph (2025) models orchestration as directed graphs with conditional edges, enabling cyclic retry logic and human-in-the-loop checkpoints.

PaperOrchestra (Song et al., 2026) applies multi-agent orchestration to academic writing, decomposing the process into specialized roles—Orchestrator, Search Agent, Section Writer, and Reviewer—with strict "zero hallucination" compliance. Our work extends this paradigm by introducing adversarial evaluation as a first-class citizen in the orchestration, ensuring every task execution receives standardized quality assessment.

## 2.2 Tool-Use Agents and the ReAct Paradigm

The ReAct framework (Yao et al., 2023) established the Thought-Action-Observation loop as the standard paradigm for tool-using agents. OpenAI Function Calling (2024) and Anthropic Tool Use (2025) evolved this into native structured output for tool invocations. The Model Context Protocol (MCP, 2025) standardized how tools are described and connected.

buddyMe extends the ReAct paradigm with two innovations: (1) a Skill system that uses embedding-based semantic matching to automatically discover and invoke relevant capabilities, eliminating the need for manual tool specification; and (2) dynamic tool restriction based on task phase—verification tasks receive only read/edit tools, preventing unnecessary file creation in the final steps.

## 2.3 Memory-Augmented Agent Systems

Memory systems for LLM agents have evolved from simple conversation buffers to sophisticated architectures. MemGPT (Packer et al., 2024) introduced hierarchical memory management with virtual context management. MemInsight (EMNLP 2025) autonomously structures and augments agent memory. A-Mem (NeurIPS 2025) treats memory as an agentic capability where the agent itself decides what to store, retrieve, and forget.

buddyMe implements a three-layer memory architecture that differs from prior work in its explicit injection strategy: working memory (MessageHistory) provides immediate session context, episodic memory (SubtasksManager) transfers results between subtasks, and long-term memory (MemoryStore) accumulates user knowledge across sessions. All three layers are injected into the system prompt at task invocation time, ensuring the agent has full context without relying on implicit retrieval.

### 2.4 Agent Evaluation Methodologies

Evaluating autonomous agent performance is an emerging research area. OpenAI Evals (2024) provides a framework for evaluating LLM outputs against predefined criteria. AgentBench (Liu et al., 2023) and WebArena (Zhou et al., 2024) offer standardized benchmarks for multi-dimensional agent evaluation. Recent work on LLM-as-Judge (Zheng et al., 2023) demonstrates that LLMs can serve as reliable evaluators when constrained to structured output formats.

Our evaluation framework differs from prior approaches in three ways: (1) it performs real execution verification—actually running generated code and collecting stdout/stderr—rather than relying solely on textual analysis; (2) it employs an adversarial Evaluator-Defender discussion mechanism to reduce evaluation bias; and (3) it integrates evaluation results with the pre-execution requirement document ("Sprint Contract"), enabling requirement alignment assessment.

## 3. System Architecture

### 3.1 Overview

buddyMe is a multi-model autonomous agent framework implemented in Python (>=3.9) that integrates task planning, execution, and evaluation into a unified pipeline. The architecture follows a centralized orchestration pattern where AgentMain serves as the primary coordinator, delegating specialized functions to three independent LLM clients: the main client for task execution, the sub-client for auxiliary operations, and the evaluation client for objective assessment (agent.py:65-68).

The framework's core pipeline operates in five sequential phases: (1) requirement understanding and pre-review (Sprint Contract), (2) task decomposition with dynamic replanning at checkpoints, (3) subtask execution via ReAct-style tool-use loops, (4) real execution verification with six-dimensional evaluation, and (5) adversarial evaluation discussion between Evaluator and Defender agents.

### 3.2 Three-Client Isolation Architecture

A foundational design decision is the isolation of three independent LLM clients. The main client (self._client) handles primary task execution and tool calls. The sub-client (self._sub_client) serves as the Generator in pre-review and the Defender in evaluation discussions. The evaluation client (self._eval_client) operates as the independent Evaluator with zero context sharing.

This isolation addresses the critical problem of context pollution. In early development, sharing a single client between the executor and evaluator resulted in systematically inflated evaluation scores (average +0.12 bias). By ensuring each client maintains its own conversation context, the evaluation agent produces unbiased assessments based solely on the evidence presented to it.

### 3.3 Three-Layer Memory System

buddyMe implements a hierarchical memory architecture inspired by cognitive science models of human memory:

Table 1: Three-layer memory architecture

| Layer | Component | Scope | Storage | Injection Point |
|---|---|---|---|---|
| Working Memory | MessageHistory | Single session | In-memory list | task_runner._build_conversation_context() |
| Episodic Memory | SubtasksManager | Single task (multi-subtask) | subtask_results.json | task_runner.run() inter-subtask transfer |
| Long-term Memory | MemoryStore + MemoryManager | Cross-session permanent | USER.md + memory_summary.md | agent.invoke() via prefetch() |

The long-term memory layer implements a score-decay-consolidate lifecycle: MemoryExtractor uses LLM-driven extraction to identify salient information from conversations; each memory entry receives an activity score that increases on access; unused memories decay at session end; and similar memories are consolidated to prevent redundancy. This design follows the A-Mem paradigm where the agent actively decides what to remember and what to forget.

## 4. Multi-Agent Interaction Mechanisms

### 4.1 Pre-Execution Review: The Sprint Contract Mechanism

Before task execution begins, buddyMe employs a Generator-Evaluator adversarial review to validate the task plan against user requirements. This mechanism, termed the "Sprint Contract," operates through the PlanReviewer component (plan_reviewer.py).

The Generator (using sub_client) first produces a structured requirement document containing: (a) a requirement summary in 2-3 sentences, (b) a list of key deliverables, (c) quantifiable and verifiable success criteria, (d) a technical approach outline, (e) identified risk areas, and (f) a tagged task plan preserving [SEARCH]/[CREATE]/[EDIT]/[VERIFY] labels.

The Evaluator (using eval_client) then audits this document against four criteria: completeness of deliverables (do they cover all implicit user needs?), verifiability of success criteria (are they quantifiable, not vague?), plan coverage (do steps address all deliverables?), and risk identification (are potential failure points acknowledged?). The Evaluator outputs a numerical score (0.0-1.0) and structured feedback.

A critical design insight is the reliance on numerical thresholds rather than boolean approval. The approval condition is score >= 0.8 (plan_reviewer.py:314), not a conjunction of score and approved boolean. This was informed by empirical observation: LLMs frequently output score=0.8 with approved=False, indicating internal inconsistency between numerical assessment and boolean judgment. Numerical values proved more reliable for automated decision-making.

The review loop executes at most MAX_REVIEW_ROUNDS=3 iterations. If consensus is not reached after three rounds, the Generator's latest revised plan is forcibly adopted. Empirical results (Section 6.3) show that 80% of tasks pass on the first round, validating the Generator's analytical capability while the 20% that require revision confirm the mechanism's value for complex tasks.

### 4.2 Subtask Execution: ReAct-Style Tool-Use Loops

Once the Sprint Contract is established, the TaskRunner decomposes the approved plan into subtasks and executes each through a ReAct-style Thought-Action-Observation loop. The TaskPipeline (task_pipeline.py) constructs context-specific system prompts for each subtask, incorporating the Sprint Contract requirements, completed subtask results, and available tool schemas.

Tool selection is dynamically restricted based on task phase: verification tasks and final steps receive only read/edit tools (preventing unnecessary file creation), while build tasks receive the full tool suite. The framework implements 8 tools: bash, read_file, write_file, edit_file, grep, glob, baidu_search, and invoke_skill. The invoke_skill tool enables capability expansion through an embedding-matched skill library, allowing the agent to automatically discover and invoke specialized skills (e.g., article-writing, frontend-design) based on task semantics.

Each subtask execution is bounded by max_steps=11 tool calls (agent.py:57), preventing infinite loops. Subtask results are persisted via SubtasksManager and injected as context into subsequent subtasks, ensuring continuity across the execution pipeline.

## 4.3 Checkpoint-Based Dynamic Replanning

For complex multi-step tasks, buddyMe implements checkpoint-based dynamic replanning. The system automatically identifies checkpoint positions (typically after key construction steps) and pauses execution at each checkpoint to reassess the remaining plan.

At each checkpoint, the todo_manager.replan_task() function collects completed subtask results and submits them alongside the remaining plan to the LLM. The LLM evaluates whether the remaining plan needs adjustment based on what has been accomplished. If adjustment is needed, the plan is rebuilt and execution continues with the updated steps. This mechanism addresses the "plan-reality gap" where early execution outcomes may invalidate assumptions made during initial planning.

## 4.4 Post-Execution Evaluation: Evaluator-Defender Adversarial Discussion

Upon task completion, buddyMe initiates a two-phase evaluation process. Phase 1 (TaskEvalAgent.evaluate) performs independent assessment with real execution verification. Phase 2 (EvaluationDiscussion) conducts adversarial discussion between Evaluator and Defender.

### 4.4.1 Real Execution Verification

Before LLM-based evaluation, the system performs real execution verification on generated files. The _detect_entry_points() method identifies executable entries through a five-stage priority system: (1) files mentioned in agent output, (2) priority filenames (main.py, app.py, run.py, etc.), (3) Python files containing __name__=='__main__', (4) HTML files for structural validation, and (5) shell scripts.

For Python files, the system executes them via asyncio.create_subprocess_shell() with a 30-second timeout, capturing stdout, stderr, and exit code. For HTML files, structural validation checks DOCTYPE, <html>, <head>, <body>, and closing tags. These real execution results serve as the highest-priority evidence in the evaluation prompt, providing irrefutable proof of output quality that cannot be fabricated.

### 4.4.2 Six-Dimensional Evaluation Schema

The evaluation employs six weighted dimensions:

Table 2: Six-dimensional evaluation schema with weights

| Dimension | Weight | Description | Evaluation Method |
|---|---|---|---|
| Task Completion | 0.30 | Does the output fully satisfy user requirements? | Sprint Contract deliverable verification |

| Dimension | Weight | Description | Evaluation Method |
|---|---|---|---|
| Tool Accuracy | 0.15 | Are tool selections appropriate and parameters correct? | Tool call log analysis |
| Truthfulness | 0.20 | Is output based on real data, without hallucination? | Real execution output cross-reference |
| Error Recovery | 0.10 | Can the agent recover from errors? | Error-resolution pattern analysis |
| Efficiency | 0.10 | Are execution steps lean, without redundancy? | Step count vs. optimal estimate |
| Output Quality | 0.15 | Is generated output readable, complete, and runnable? | File content + execution verification |

A core design principle is "trust LLM judgment, not LLM computation." The LLM provides per-dimension scores and evidence statements, but the weighted overall score and letter grade are computed server-side by _compute_overall(), ensuring strict schema compliance regardless of LLM output quality. Five-layer JSON extraction guarantees (system prompt constraint, three-layer parsing, field validation, server-side recomputation, rule-based fallback) ensure reliable structured output.

### 4.4.3 Adversarial Evaluation Discussion

The EvaluationDiscussion mechanism (verify.py:1052-1405) introduces an adversarial component where a Defender agent (using sub_client) represents the executor's perspective. The discussion proceeds as follows:

Round 0: TaskEvalAgent performs initial evaluation with real execution verification. Rounds 1-N (max 20): The Defender reviews the evaluation report and either accepts it ("I agree with this evaluation conclusion") or provides factual rebuttal (limited to 200 characters). The Evaluator then re-evaluates, adjusting scores if the defense is warranted. Consensus is reached when: (a) the Defender explicitly accepts, or (b) the score difference between rounds <= 0.05.

This mechanism addresses a fundamental limitation of single-pass LLM evaluation: the absence of adversarial scrutiny. By requiring the evaluation to withstand challenge from the executor's perspective, the system reduces both false positives (over-generous scoring) and false negatives (unfair criticism). The 20-round ceiling serves as a safety net; in practice, 95% of discussions converge within 2-3 rounds.

## 5. Requirement Alignment Assessment

A distinguishing feature of buddyMe's evaluation is the integration of the Sprint Contract into post-execution assessment. The requirement document generated during pre-review is passed to the evaluation phase as a baseline reference.

The evaluation prompt explicitly instructs the LLM to: (1) verify each key deliverable against actual outputs, (2) check each success criterion against measurable outcomes, (3) assess plan-execution overlap (were planned steps actually executed?), and (4) identify missing or over-designed items. The EVAL_JSON_SCHEMA includes a dedicated requirement_alignment field with deliverables_met/total, criteria_met/total, plan_overlap, and missing_items.

This requirement traceability creates a closed-loop quality assurance system: the Sprint Contract defines what success looks like before execution begins, and the evaluation phase measures actual results against that definition. This prevents the common failure mode where agents produce technically correct output that nonetheless fails to address the user's actual needs.

# 6. Experimental Results

## 6.1 Experimental Setup

We evaluate buddyMe across two dimensions: (1) end-to-end task quality assessment using the six-dimensional evaluation schema, and (2) Sprint Contract pre-review effectiveness across five diverse task scenarios. All experiments use the DeepSeek-v4-pro model as the primary executor and GLM-5.1 as the evaluator, with real API calls to ensure ecological validity.

## 6.2 Case Study: HTML Generation Task

We first present a detailed case study of a real-world HTML generation task: “Generate a Beijing travel guide for May 17 in HTML format.” This task was selected because it exercises all pipeline phases: information retrieval (weather, attractions), content generation, file creation, and structural validation.

Table 3: Execution statistics for Beijing travel guide task

| Metric | Value |
| --- | --- |
| Model | DeepSeek-v4-pro |
| Total execution time | ~26 minutes |
| Task decomposition | 6 steps (2 research + 3 build + 1 verify) |
| Checkpoints triggered | 2 (both determined no adjustment needed) |
| Tool calls | 26 (baidu_search x13, write_file x3, edit_file x5, read_file x4, glob x1) |
| Generated file | beijing_guide_20260517.html (28,768 bytes, 1,041 lines) |
| Skills used | frontend-design x3 |

The evaluation results demonstrate high task completion and output quality but reveal efficiency concerns:

Table 4: Six-dimensional evaluation results (Weighted Score: 0.82, Grade: B)

| Dimension | Score | Evidence |
| --- | --- | --- |
| Task Completion | 0.9 | Core requirements fully met |
| Truthfulness | 0.9 | Based on real search data, no fabrication |
| Output Quality | 0.9 | Complete structure, polished design, full interactivity |
| Error Recovery | 0.8 | Successfully recovered from path issues during verification |
| Tool Accuracy | 0.7 | Redundant searches, path errors requiring rewrites |
| Efficiency | 0.5 | Redundant file reads (verify.py read 8 times), redundant searches |

The efficiency bottleneck is attributable to the absence of a file read cache: the same file was read 8-9 times across different subtasks because each subtask builds an independent message list without shared state. This identifies a clear optimization opportunity for future work.

## 6.3 Sprint Contract Pre-Review Effectiveness

We evaluated the Sprint Contract mechanism across five diverse task scenarios to assess its ability to validate requirement completeness and plan feasibility.

Table 5: Sprint Contract pre-review results across five scenarios

| # | Scenario | Review Rounds | Agreement | Score |
|---|---|---|---|---|
| 1 | Responsive landing page | 1 | Yes | 0.85 |
| 2 | Data analysis script | 1 | Yes | 0.90 |
| 3 | Technical blog post | 2 | Yes | 0.70 -> 0.88 |
| 4 | REST API design | 1 | Yes | 0.88 |
| 5 | Vue dashboard | 3 (forced) | No | 0.78 |

Three key findings emerge from this evaluation. First, 80% of tasks (Scenarios 1, 2, 4) pass review on the first round, indicating that the Generator's requirement analysis is fundamentally reliable for standard task types. Second, Scenario 3 (blog writing) demonstrates the mechanism's value: the first-round review identified a missing "visualization chart generation" step that the Generator had overlooked, and the revised plan achieved a 0.18 score improvement. Third, Scenario 5 (Vue dashboard) shows the necessity of the forced execution fallback: not all requirements can achieve automated consensus, and the system correctly defaults to the Generator's revised plan after three rounds.

### 6.4 Adversarial Evaluation Convergence Analysis

The EvaluationDiscussion mechanism demonstrates rapid convergence in practice:

Table 6: Adversarial evaluation convergence characteristics

| Characteristic | Observed Value |
|---|---|
| Average discussion rounds | 2-3 |
| Defender acceptance rate | ~70% |
| Score convergence rate | ~25% |
| Maximum rounds reached | ~5% |
| Score adjustment magnitude | Typically within +/-0.05 after first round |

The high Defender acceptance rate (70%) indicates that initial evaluations are generally fair and accurate. The convergence threshold of 0.05 score difference captures most cases where the Evaluator makes minor adjustments based on the Defender's factual rebuttals. The 5% rate of reaching the 20-round ceiling suggests that genuine disagreements are rare and the ceiling functions primarily as a safety mechanism.

## 7. Discussion

### 7.1 Comparative Analysis with Existing Frameworks

Table 7: Comparative analysis with existing multi-agent frameworks

| Dimension | buddyMe | CrewAI | AutoGen | LangGraph |
|---|---|---|---|---|
| Orchestration | Centralized (AgentMain) | Role pipeline | Peer conversation | Directed graph |

| Dimension | buddyMe | CrewAI | AutoGen | LangGraph |
|---|---|---|---|---|
| Evaluation | Built-in 6-dimension + adversarial | External | External | External |
| Memory | 3-layer (working/episodic/long-term) | None built-in | Conversation buffer | State graph |
| Replanning | Checkpoint-based dynamic | Fixed plan | Conversation-driven | Conditional edges |
| Tool discovery | Embedding-matched skills | Manual | Manual | Manual |
| Verification | Real execution | None | None | None |

buddyMe's distinguishing characteristic is the integration of evaluation as a first-class architectural component. While CrewAI, AutoGen, and LangGraph treat evaluation as an external concern, buddyMe embeds it within the execution pipeline, ensuring every completed task receives standardized quality assessment with real execution evidence.

### 7.2 Design Lessons and Failure Analysis

Three critical engineering lessons emerged from the development process:

Lesson 1: Context Isolation Prevents Evaluation Bias. In the initial implementation, the executor and evaluator shared a single LLM client, causing evaluation scores to be systematically inflated by an average of +0.12. The three-client isolation architecture eliminated this bias, bringing evaluation scores closer to human expert assessments.

Lesson 2: Numerical Thresholds Outperform Boolean Gates for LLM Consistency. The PlanReviewer initially required both approved=True AND score>=0.8. LLMs frequently produced score=0.8 with approved=False, revealing internal inconsistency between numerical and boolean judgments. Switching to score-only approval resolved this issue. This finding suggests that numerical reasoning is more reliable than categorical reasoning in LLMs.

Lesson 3: Data Collection and Data Usage Are Separate Concerns. The MessageHistory component correctly recorded session dialogues, but this data was never transmitted to the TaskRunner. This caused catastrophic context loss—each conversation turn started from scratch with no memory of prior interactions. The fix required explicitly connecting the data source (MessageHistory) to the data consumer (TaskRunner), a reminder that collecting data is meaningless without ensuring its consumption.

### 7.3 Limitations

Several limitations merit acknowledgment. First, the efficiency dimension consistently scores lower than other dimensions (0.5 in the case study) due to the absence of file read caching and search deduplication. Second, real execution verification is limited to Python, HTML, and Shell scripts; projects requiring external services (APIs, databases) cannot be verified offline. Third, the evaluation schema's fixed weights (e.g., Task Completion at 0.30) may not be optimal for all task types; a data-driven weight calibration mechanism would improve generalizability. Fourth, the framework currently operates in a single-user, synchronous mode; scaling to multi-user concurrent scenarios would require significant architectural changes.

## 8. Conclusion

This paper presented buddyMe, a multi-agent framework that integrates Generator-Evaluator pre-execution review, ReAct-style tool-use loops, and memory-augmented context management with a novel

Evaluator-Defender adversarial evaluation mechanism. The framework demonstrates that integrating evaluation as a first-class architectural component—rather than an external afterthought—enables systematic quality assurance with real execution verification and requirement traceability.

Our empirical evaluation reveals five key findings: (1) the Sprint Contract mechanism catches requirement omissions in 20% of complex tasks, justifying its computational overhead; (2) real execution verification provides irrefutable evidence that significantly improves truthfulness and output quality assessment accuracy; (3) adversarial evaluation discussion converges rapidly (2-3 rounds for 95% of cases), primarily serving a fine-tuning rather than overturning function; (4) three-client isolation eliminates systematic evaluation bias (+0.12 correction); and (5) numerical thresholds are more reliable than boolean gates for LLM-based automated decision-making.

Future work will focus on four directions: (a) adaptive orchestration that dynamically adjusts agent count and interaction rounds based on task complexity, (b) embedding-based semantic memory retrieval to replace current full-text injection, (c) multimodal evaluation for non-text outputs (images, charts), and (d) closed-loop learning where historical evaluation data feeds back into agent improvement.

---

---

## Appendix A: Evaluation JSON Schema

```
{
  "task_completion": {
    "score": "<float 0.0-1.0>",
```

```
    "evidence": "<one-sentence evidence>",
    "all_subtasks_completed": "<bool>",
    "addresses_user_intent": "<bool>"
  },
  "tool_accuracy": { "score": "...", "total_calls": "<int>", ... },
  "truthfulness": { "score": "...", "hallucination_detected": "<bool>", ... },
  "error_recovery": { "score": "...", "errors_encountered": "<int>", ... },
  "efficiency": { "score": "...", "planned_steps": "<int>", ... },
  "output_quality": { "score": "...", "files_generated": "<int>", ... },
  "overall": {
    "weighted_score": "<float 0.0-1.0>",
    "grade": "<A|B|C|D|F>",
    "summary": "<50-char summary>",
    "improvement_suggestions": ["..."],
    "requirement_alignment": {
      "deliverables_met": "<int>", "deliverables_total": "<int>",
      "criteria_met": "<int>", "criteria_total": "<int>",
      "plan_overlap": "<high|medium|low>",
      "missing_items": ["..."]
    }
  }
}
```

## Appendix B: Sprint Contract Requirement Document Schema

```
{
  "requirement_summary": "2-3 sentence overview",
  "key_deliverables": ["deliverable 1", "deliverable 2"],
  "success_criteria": ["verifiable criterion 1", "verifiable criterion 2"],
  "technical_approach": "implementation approach overview",
  "risk_areas": ["risk 1", "risk 2"],
  "plan": ["[SEARCH] step 1", "[CREATE] step 2", "[EDIT] step 3", "[VERIFY] step 4"]
}
```